\pdfoutput=1

\documentclass[11pt]{article}

\usepackage[]{acl}

\usepackage{times}
\usepackage{latexsym}

\usepackage[T1]{fontenc}

\usepackage[utf8]{inputenc}

\usepackage{times}
\usepackage{latexsym}
\usepackage{cancel}
\usepackage{graphicx}
\usepackage{amsmath}
\usepackage{latexsym}
\usepackage{amssymb}
\usepackage{amsthm}
\usepackage{color}
\usepackage{xspace}
\usepackage{subfigure}
\usepackage{array, booktabs, makecell}

\usepackage{booktabs}
\usepackage{array,tabularx}
\usepackage[ruled,vlined]{algorithm2e}
\usepackage{algpseudocode}
\usepackage{ulem}
\usepackage{amssymb}
\usepackage{pifont}
\usepackage{multirow}

\title{Template Filling for Controllable Commonsense Reasoning}

\author{Dheeraj Rajagopal$^\spadesuit$, \textbf{Vivek Khetan$^\clubsuit$}, \textbf{Bogdan Sacaleanu$^\clubsuit$}, \\ 
\textbf{Anatole Gershman}$^\spadesuit$  
\textbf{Andrew Fano$^\clubsuit$}, \textbf{Eduard Hovy$^\spadesuit$} \\
  $^\spadesuit$ Language Technologies Institute, Carnegie Mellon University, Pittsburgh, PA, USA \\ 
  $^\clubsuit$ Accenture Labs, San Francisco, USA \\ 
  \texttt{\{dheeraj, anatoleg, hovy\}@cs.cmu.edu} \\ \texttt{\{vivek.a.khetan, andrew.e.fano,  bogdan.e.sacaleanu\}@accenture.com} \\} 

\date{}
\begin{document}
\definecolor{Red}{rgb}{1,0,0}
\definecolor{Green}{rgb}{0,1,0}
\definecolor{Blue}{rgb}{0,0,1}
\definecolor{Red}{rgb}{0.9,0,0}
\definecolor{Orange}{rgb}{1,0.5,0}
\definecolor{yellow}{rgb}{0.65,0.6,0}
\definecolor{cadmiumgreen}{rgb}{0.2, 0.7, 0.24}

\newcommand{\amn}[1]{\textcolor{yellow}{[#1 \textsc{--Aman}]}}
\newcommand{\vivek}[1]{\textcolor{Red}{[#1 \textsc{--vivek}]}}
\newcommand{\dheeraj}[1]{\textcolor{Orange}{[#1 \textsc{--Dheeraj}]}}
\newcommand{\ed}[1]{\textcolor{Blue}{[#1 \textsc{--Ed}]}}
\newcommand{\secref}[1]{\S\ref{#1}}
\newcommand\given[1][]{\:#1\vert\:}

\newcommand{\V}[1]{\mathbf{#1}}
\newcommand{\eat}[1]{}

\newcommand{\numgraphs}{2107\xspace}
\newcommand{\numpassages}{379\xspace}
\newcommand{\numquestions}{40.7K\xspace}
\newcommand{\grn}[1]{\textcolor{cadmiumgreen}{#1}}
\newcommand{\red}[1]{\textcolor{Red}{#1}}

\newcommand{\lrate}{\textcolor{Red}{LR-HERE} }
\newcommand{\dropout}{\textcolor{Red}{DROPOUT-HERE} }
\newcommand{\green}[1]{\textcolor{green}{#1}}
\newcommand{\cadmiumgreen}[1]{\textcolor{cadmiumgreen}{#1}}

\newcommand{\helps}{\overset{+}{\longrightarrow}}
\newcommand{\hurts}{\overset{-}{\longrightarrow}}
\newcommand{\helpedby}{\overset{+}{\longleftarrow}}
\newcommand{\hurtby}{\overset{-}{\longleftarrow}}
\newcommand{\entails}{\overset{\oplus}{\longrightarrow}}
\newcommand{\doesnotentail}{\overset{\oplus}{\longleftarrow}}
\newcommand{\relatedby}{\overset{r}{\longrightarrow}}
\newcommand{\mask}{\texttt{[MASK]}\xspace}
\newcommand{\blank}{\texttt{[BLANK]}\xspace}
\newcommand{\prompt}{\texttt{PROMPT}\xspace}
\newcommand{\ourmodel}{\texttt{POTTER}\xspace}
\newcommand{\approach}{\text{TemplateCSR}\xspace}
\newcommand{\csk}{\textit{commonsense}\xspace}
\newcommand{\health}{\textit{health and well-being}\xspace}
\newcommand{\template}{\texttt{SPL TOKEN}\xspace}

\newcommand{\bert}{\textsc{BERT}\xspace}
\newcommand{\bart}{\textsc{BART}\xspace}
\newcommand{\tfive}{\textsc{T5}\xspace}
\newcommand{\bartbase}{\textsc{BART-Base}\xspace}
\newcommand{\bartlarge}{\textsc{BART-Large}\xspace}
\newcommand{\tfivebase}{\textsc{T5-Base}\xspace}
\newcommand{\tfivelarge}{\textsc{T5-Large}\xspace}
\newcommand{\bertbase}{\textsc{BERT-Base}\xspace}
\newcommand{\bertlarge}{\textsc{BERT-Large}\xspace}


\newcommand{\activity}[1]{\textcolor{teal}{\texttt{\textbf{#1}}}}
\newcommand{\reason}[1]{\textcolor{red}{\texttt{\textbf{#1}}}}
\newcommand{\disease}[1]{\textcolor{purple}{\texttt{\textbf{#1}}}}
\newcommand{\qual}[1]{\textcolor{cyan}{\texttt{\textbf{#1}}}}

\newcommand{\plm}{\textsc{seq-to-seq}\xspace}
\newcommand{\corr}[2]{\textbf{\textcolor{red}{\sout{#1} #2}}}
\newcommand{\rouge}{\texttt{ROUGE}\xspace}
\newcommand{\bertscore}{\texttt{BERTSCORE}\xspace}
\newcommand{\factcc}{\texttt{FACTCC}\xspace}
\newcommand{\dac}{\texttt{DAE}\xspace}

\newcommand{\squishlist}{
  \begin{list}{$\bullet$}
    { \setlength{\itemsep}{0pt}      \setlength{\parsep}{3pt}
      \setlength{\topsep}{3pt}       \setlength{\partopsep}{0pt}
      \setlength{\leftmargin}{1.5em} \setlength{\labelwidth}{1em}
      \setlength{\labelsep}{0.5em} } }
\newcommand{\reallysquishlist}{
  \begin{list}{$\bullet$}
    { \setlength{\itemsep}{0pt}    \setlength{\parsep}{0pt}
      \setlength{\topsep}{0pt}     \setlength{\partopsep}{0pt}
      \setlength{\leftmargin}{0.2em} \setlength{\labelwidth}{0.2em}
      \setlength{\labelsep}{0.2em} } }

 \newcommand{\squishend}{
     \end{list} 
 }

\maketitle
\begin{abstract}

Large-scale sequence-to-sequence models have shown to be adept at both multiple-choice and open-domain commonsense reasoning tasks. However, the current systems do not provide the ability to control the various attributes of the reasoning chain. To enable better controllability, we propose to study the \textit{commonsense reasoning as a template filling task} (\approach) --- where the language models fills reasoning templates with the given constraints as control factors.
As an approach to \approach, we (i) propose a dataset of commonsense reasoning template-expansion pairs and (ii) introduce \ourmodel, a pretrained sequence-to-sequence model using prompts to perform commonsense reasoning across concepts. Our experiments show that our approach outperforms baselines both in generation metrics and factuality metrics. We also present a detailed error analysis on  our approach's ability to reliably perform commonsense reasoning\footnote{All code and data will be released publicly}.

\end{abstract}

\section{Introduction}

Commonsense reasoning has been studied across both \textit{multiple choice} \citep{tandon-etal-2018-reasoning,talmor2019commonsenseqa,Lv2020GraphBasedRO} and \textit{open-ended knowledge base} settings \citep{Lin2021DifferentiableOC}. While multiple choice approaches require a list of answer options, open-ended KB approaches assume that the answer exists in an available knowledge base (KB). Such constraints often limit these systems' ability in practical applications where control is required (e.g. a web search query with specific conditions). 

\begin{figure}[!t]
    \centering
    \includegraphics[width=0.9\linewidth]{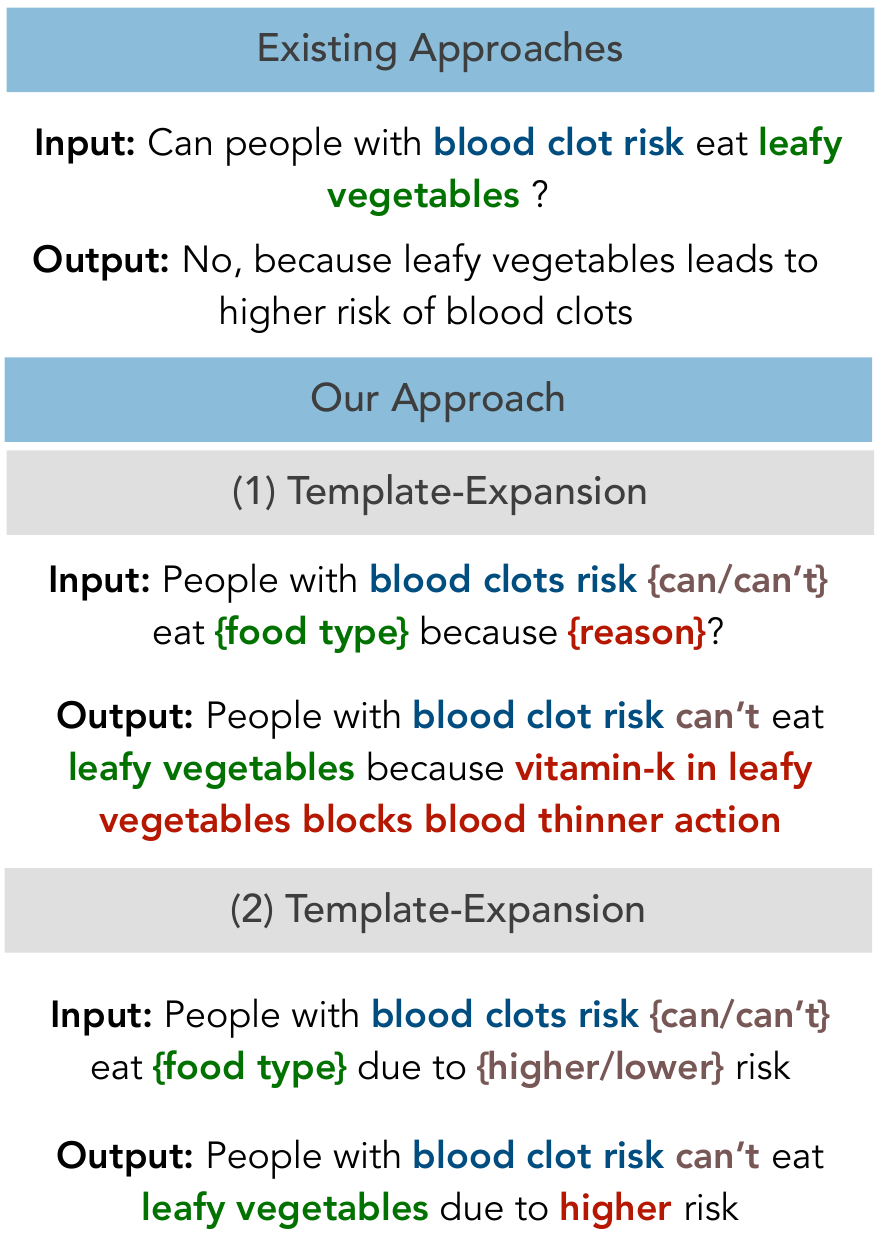}
    \caption{In this example, we show how a commonsense reasoning question can be formulated as two different template-expansion pairs, each focusing on different aspects of reasoning between the concepts \textit{smoking} and \textit{lung cancer}. While formulation (1) focuses on the explanation, (2) aims to understand the qualitative relationship between them. }
    \label{fig:intro_example}
\end{figure}

To complement the existing commonsense reasoning efforts, our work aims to enhance the commonsense reasoning capabilities of natural language processing (NLP) systems by studying \textbf{template commonsense reasoning} (TemplateCSR) --- where reasoning is achieved by filling templates with restricted template slots, rather than selecting answers from a list of candidates or KB. TemplateCSR task is challenging as there are no available annotations and potentially multiple correct expansions for each template.
Moreover, the task of designing templates with slots that satisfy arbitrary constrains is still an open challenge.
For example, for an example reasoning template \textbf{People who smoke are at a risk of \texttt{\{disease\}} }, a system needs to first constrain the slot to only \textit{diseases}, and then use the additional constraint of \textit{smoking} to arrive at the right answer in the slot. In comparison to Language Model (LM) probing approaches \citep{ribeiro-etal-2020-beyond} that test capabilities of LM that are already trained, we aim to propose a model for \approach task.

\begin{figure*}[!th]
\centering
\begin{minipage}{1.0\textwidth}
  \centering
  \includegraphics[width=\linewidth,keepaspectratio]{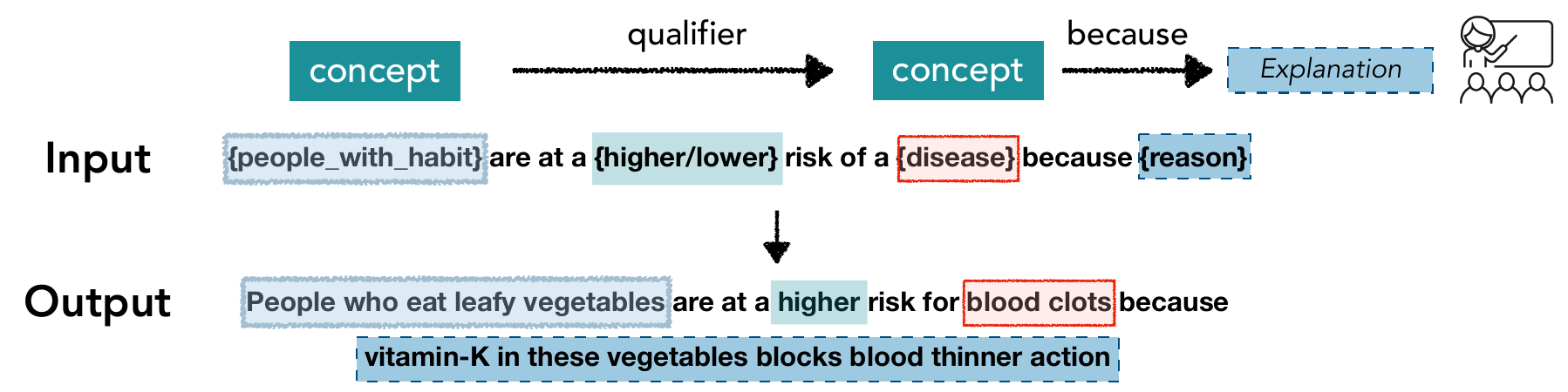}
  \caption{An overview of the overall template structure for our approach. Our goal is to reason across concepts for \approach. In this example template, concept slots are \textit{people\_with\_habit} and \textit{disease}, and the multiple choice qualifier slot - \textit{higher/lower} describes their relationship and an explanation \textit{reason} slot aims to get a free-form text explanation for how they are related.}
  \label{fig:approach_example}
\end{minipage}%

\end{figure*}

Figure \ref{fig:intro_example} shows one such example, where we show how an existing commonsense reasoning query can be formulated as different template-expansion pairs with control over different aspects of the reasoning. In the first expansion, the reasoning chain focuses on the relationship between \textit{smoking} and \textit{cancer} with the corresponding explanation (\texttt{reason}), while the second chain solely focuses on the qualitative relationship between the \textit{smoking} and \textit{cancer}. 

To address the above mentioned challenges, our contributions in this paper for \approach are two-fold. First, we present a dataset of commonsense reasoning templates and their corresponding expansions that are valid completions of the template, which we define as template-expansion pairs \citep{Fass1983PreferenceSI}. 
The slots in the templates are open-ended and are not restricted to any particular categories and enable controlling the reasoning chain. Given the recent focus on explainable models for reasoning \citep{Wiegreffe2021TeachMT}, we also augment templates with an optional free-form explanation slot that explains the reasoning connection between various commonsense concepts. Our \approach dataset comprises of about 3600 unique template-expansion pairs collected from diverse sources, and we hope to enable \plm systems to effectively learn to fill commonsense reasoning templates. 

Next, we present \ourmodel, a model that formulates the \approach challenge as a \plm task where given a template with slots for specific concepts, the goal of the model is to produce meaningful completed sentences for the template. The concept in each slot in the template is provided via a \textit{prompt} \citep{gpt3-NEURIPS2020_1457c0d6}, which indicates an abstraction of the nature of the slot. The multiple choice \textit{qualifier} slot helps model the relationship between the concepts and the \textit{explanation} slot generates a free-form text explanation for the reasoning chain. Specifying each slots in free-form text enables control allowing commonsense reasoning questions to specify concepts, the qualitative relationship and the nature of explanation. 

In our experiments for the \approach task, \ourmodel outperforms baseline  both in terms of generation metrics such as \rouge and \bertscore, and factual correctness (factuality) metrics such as \factcc. We also evaluate the factuality using human judges and a detailed analysis of model outputs. While we still observe factual errors, our approach provides a more nuanced understanding of the mistakes, potentially expanding the way commonsense reasoning systems can be built using \plm models. 


\section{Dataset} 

For our use-case, we create dataset samples of commonsense reasoning templates related to \textit{lifestyle and health}. Incorporating NLP systems for aiding healthy lifestyle has been an active area of research in the past decade \citep{Liberato2014NutritionIA, Fadhil2017AddressingCI,Doustmohammadian2021SocialMI, Ahne2022ExtractionOE}. Inspired by this line of research, we want to collect templates that describe a relation between lifestyle related commonsense concept and a corresponding health related concept. In comparison to existing datasets like \textit{commonsenseqa} \citep{talmor2019commonsenseqa} which relies on fixed set of relationships from a knowledge-base \citep{Speer2013ConceptNet5A}, we do not restrict the relationship types or number of concepts or hops, making it close to open-vocabulary text. We believe that our dataset augments well with the existing commonsense reasoning datasets in the community, contributing to the diversity of the data. 

\begin{table*}[t]
\centering
\resizebox{\textwidth}{!}{%
\begin{tabular}{@{}lll@{}}
\toprule
\multirow{2}{*}{Concepts} &
  \multicolumn{2}{c}{Template Expansion Pairs} \\ \cmidrule(l){2-3} 
 &
  Sample Template &
  Valid Expansions for the Template \\ \midrule
\begin{tabular}[c]{@{}l@{}}location, \\ disease\end{tabular} &
  \begin{tabular}[c]{@{}l@{}}\activity{\{person\_at\_location\}} has a\\ \qual{\{higher/lower\}} risk of \activity{\{disease\}} \\ because  \reason{\{reason\_for\_risk\}}\end{tabular} &
  \begin{tabular}[c]{@{}l@{}}Person who lives in a city has a higher risk \\ - of depression because of stress due to noise \\ Person who lives near a village has a lower risk \\ - of respiratory illness because of lower pollution\end{tabular} \\ \midrule
\begin{tabular}[c]{@{}l@{}}prescription\\ medication, \\ disease\end{tabular} &
  \begin{tabular}[c]{@{}l@{}}\activity{\{person\_taking\_prescription\}}\\ has a higher risk\\ of \activity{\{disease\}} \\ due to \reason{\{reason\}}\end{tabular} &
  \begin{tabular}[c]{@{}l@{}}Someone on steroids have a higher risk for heart disease\\ - because steroids compromise heart pumping\\ People on insulin have a lower risk of hyperglycemia \\ - because of lower glucose levels.\end{tabular} \\ \midrule
\begin{tabular}[c]{@{}l@{}}food item, \\ substitute \\ item\end{tabular} &
  \begin{tabular}[c]{@{}l@{}}\activity{\{food\_item\_1\}} should not \\ be consumed with \activity{\{food\_item\_2\}} \\ because \reason{\{reason\}}\end{tabular} &
  \begin{tabular}[c]{@{}l@{}}Steak should not be consumed with mashed potatoes \\ - because pairing fried foods increases the risk of diabetes.\\ Pizza should not be consumed with French fries because\\ - proteins require a much different stomach\\ - environment than starches for proper digestion\end{tabular} \\ \midrule
\begin{tabular}[c]{@{}l@{}}behavior \\ change, \\ medical \\ condition\end{tabular} &
  \begin{tabular}[c]{@{}l@{}}A change in behavior such as\\  \activity{\{behavior\_change\}} is often \\ associated with \activity{\{a\_medical\_condition\}} \\ because \reason{\{reason\_for\_condition\}}\end{tabular} &
  \begin{tabular}[c]{@{}l@{}}A change in behavior such as becoming more sedentary is \\ - often associated with obesity  \\ - because less activity leads to less calorie burning.\\ A change in behavior such as no longer drinking coffee \\ - is often associated with diminished insomnia \\ - because less caffeine equals improved sleep.\end{tabular} \\ \midrule
\begin{tabular}[c]{@{}l@{}}symptom, \\ medical \\ condition, \\ everyday \\ action\end{tabular} &
  \begin{tabular}[c]{@{}l@{}}When severe symptoms like\\ \activity{\{a\_symptom\}} for a \\ \activity{\{a\_medical\_condition\}} shows up,\\ immediately one should perform \activity{\{an\_action\}}\end{tabular} &
  \begin{tabular}[c]{@{}l@{}}When severe symptoms like confusion or disorientation \\ - for heatstroke show up, immediately, one should perform \\ - cooling actions, such as applying cooling towels.\\ When severe symptoms like unconsciousness for a \\ - heart attack show up, immediately one should\\ - call 911 and perform CPR while awaiting help.\end{tabular} \\ \midrule
\begin{tabular}[c]{@{}l@{}}lifestyle \\ activity,\\ disease\end{tabular} &
  \begin{tabular}[c]{@{}l@{}}People often do \activity{\{an\_activity\}} \\ before going to  bed in night to prevent risk \\ of \activity{\{disease\}}.  \\ This is because \reason{\{reason\_for\_activity\}}\end{tabular} &
  \begin{tabular}[c]{@{}l@{}}People often do reading before going to bed in night\\ - to prevent risk of insomnia. This is because\\ - doing some light reading helps lull you to sleep.\\ People often do teeth brushing before going to bed in night \\ - to prevent risk  of tooth decay. This is because\\ - brushing removes cavity-causing plaque from teeth.\end{tabular} \\ \bottomrule
\end{tabular}%
}
\caption{Examples from our dataset. Each template has two corresponding sentences. \activity{\texttt{[concept]}} is a concept, and \reason{[text]} represents the explanation and \qual{[text]} represents a qualifier. We show two sentences each for a template. Each template slot is given in free-form text without any restriction in vocabulary.}
\label{tab:examples_dataset}
\end{table*}

Based on the efficacy assessment for NLP systems in health and lifestyle related settings \citep{Laranjo2018ConversationalAI, Abdalrazaq2020EffectivenessAS,Hoermann2017ApplicationOS}, we designed our basic template structure. Our basic units for the \approach task is as follows: 

\begin{enumerate}
    \item \textit{concept slot} : contains an abstract category of a concept. The concept's abstraction is provided in a natural language format in open-vocabulary, without fixed class constraints. In the example shown in figure \ref{fig:approach_example}, \textit{people with habit} and \textit{disease} are concept slots. 
    
    \item \textit{multiple-choice qualifier slot} : a word or phrase that describes the nature of the relationship between the concepts. This slot is typically framed as a multiple-choice slot, where the goal is to pick an option from the choices rather than replacing the text in the template slot. Figure \ref{fig:approach_example} shows an example where the slot \textit{higher/lower} is one such multiple-choice qualifier slot. 
    
    \item \textit{explanation slot} : this optional field consists of a free-form explanation that explains the reasoning between concepts, typically marked as \textit{reason} slot.
\end{enumerate}


Towards this, we collect a set of template ($x$) and its corresponding expansions ($y$) based on this overall schema of for commonsense reasoning. 
In the example shown in figure \ref{fig:approach_example}, the template comprises of two concept slots,  (\textit{people with habit} and \textit{disease}. 
The qualifier slot (\textit{higher/lower}) specifies how one concept is connected to another concept in terms of their qualitative relationship. 
The template also includes an optional \textit{explanation} slot that specifies in free-form text how leafy vegetable intake is connected to blood clots. A valid output for the above-mentioned template is for instance, \textit{people who smoke are at a higher risk for lung cancer because carcinogens in smoke causes DNA damage}, where \textit{people with habit} is replaced by \textit{people who smoke}, and the multiple choice qualifier slot \textit{higher/lower} is replaced by \textit{higher}, and \textit{disease} slot replaced by \textit{lung cancer} and finally the \textit{reason} slot replaced by explanation of the qualitative relationship \textit{carcinogens in smoke causes DNA damage}. In this example, we show how both the template-expansion pairs aim to uncover the relationship between \textit{smoke} and \textit{lung cancer}, while also providing the flexibility to additionally constrain the reasoning chain in any way. 

\begin{table*}[t]
\centering
\begin{tabular}{@{}ll@{}}
\toprule
 Input (Template) & Output (Expansion)                                           \\ \midrule
  \begin{tabular}[c]{@{}l@{}}
The first blank is \activity{person\_at\_location}. \\ The second blank is \qual{higher/lower}. \\ The third blank is \activity{disease}. \\ The fourth blank is a \reason{reason\_for\_risk}. \\
\activity{\mask} has a  \qual{\mask} risk of \\ \activity{\mask} because \reason{\mask} \end{tabular} & \begin{tabular}[c]{@{}l@{}}Person who lives in a city \\ has a higher risk of depression \\  because of higher stress due to noise \end{tabular} \\
\bottomrule
\end{tabular}
\caption{Overview of \ourmodel approach for \approach. Each concept category is given as a prompt to the input and the slots are represented via the \mask token. The prompt describes each slot's abstraction and the task is to generate the \textit{output}. }
\label{tab:task_setup}
\end{table*}

\paragraph{Task Setup :} 

To collect our dataset using crowdsourcing, we use amazon mechanical turk platform \footnote{\url{https://www.mturk.com/}}. 
Each datapoint took $\sim$120 seconds to annotate, and we paid an average of \textdollar15 per hour. Additionally, we used a filtering step to select master annotators with an approval rate of more than 90\%. 
All the turkers were given specific instructions to input only factual information and not opinionated statements. Specifically, the turkers were instructed to use the following sources: \textit{CDC\footnote{\url{https://www.cdc.gov/}}, WebMD\footnote{\url{https://www.webmd.com/}}, Healthline\footnote{\url{https://www.healthline.com/}}} and \textit{Mayo Clinic\footnote{\url{https://www.mayoclinic.org/}}}.
The annotators were also instructed to give a template, and at least two corresponding sentences that matches the template. The statistics of the data are as follows: the average sentence length was about \textit{14.57} words, with mean 2.4 slots per template. 
Some qualitative examples from the dataset are given in the table \ref{tab:examples_dataset}. Overall, our dataset contains about 7000 template-sentence pairs with about 3600 unique templates. 
Once the templates are collected, the authors post-process the data to verify each template-expansion pair for correctness and validate that we do not have any identifying information like proper names. We then create a standard 70/10/20 train/val/test split.

\section{Model} 

Early NLP systems have often relied on rule-based templatic systems \citep{Riloff1996AutomaticallyGE,Brin1998ExtractingPA,Agichtein1999ExtractingRF,Craven2000LearningTC} due to their simplistic nature. Compared to machine learning methods, they were often rigid \citep{Yih1997TemplatebasedIE}. Despite their rigidity, template based systems are often easy to comprehend, and lend themselves to easily incorporate domain knowledge \citep{chiticariu-etal-2013-rule}. Our goal is to combine the strengths of both template-based systems and recent advances in pretrained \plm models for the task of commonsense reasoning via template expansion. 

In this work, we present \ourmodel (\textbf{P}r\textbf{o}mp\textbf{t} \textbf{Te}mplate Filling for Commonsense \textbf{R}easoning), an approach that models the \approach task as a prompt-tuning task inspired by the recent advances in prompt-tuning. Prompt-based approaches have achieved state-of-the-art performance in several few-shot learning experiments \citep{gpt3-NEURIPS2020_1457c0d6, gao-etal-2021-making, le-scao-rush-2021-many}. We aim to leverage this for the \approach task. 


Table \ref{tab:task_setup} shows an example of our task setup for our \ourmodel approach.  
In comparison to approaches such as \citet{Donahue2020EnablingLM}, our approach does not strictly enforce that that sentences only fill missing spans of text. Rather, the expanded sentences are allowed to have additional modifications. For instance, for the following input template - \activity{\{person\_at\_location\}} has a   \qual{\{higher/lower\}} risk of  \activity{\{disease\}} because \reason{\{reason\_for\_risk\}}, a valid expansion is \textit{person who lives in the city has a higher risk of depression due to noise}. 

\subsection{Training}

Given a template $x \in \mathcal{X}$ and its corresponding expansion $y \in \mathcal{Y}$, we can train any sequence-to-sequence model that models $p_\theta (y \vert x)$. Towards this, we use a pretrained sequence-to-sequence model $\mathcal{M}$ to estimate the filled template $y$ for an input $x$. 
We model the conditional distribution  $p_{\theta}(y \given x)$  parameterized by $\theta$ as $$ p_{\theta}(y \given x) =  \prod_{k=1}^{M} p_{\theta} (y^k \given x, y^{1},.., y^{k-1}) $$ where $M$ is the length of $y$.

\subsection{Inference to Decode Template Expansions}
\label{subsec:inference}

The auto-regressive factorization of \plm $p_{\theta}$ allows us to effectively cast the constrained decoding of filling the template as generating the sequence given the input $x$.
For each expansion, we sample $y_j^{1} \sim p_\theta(y \given \mathbf{x}_j)$.
Consequently, we sample $y_j^{2} \sim p_\theta(y  \given \mathbf{x}_j, y_j^{1})$, and the token generation process is repeated until we reach the end-symbol. For each symbol, the model has to decide between generating a token to replace the template slot or generate part of the template, while also ensuring the overall generated output sequence is consistent with the constraints given in the template.

\section{Experiments} 

\begin{table*}[t]
\centering
\begin{tabular}{@{}llllll@{}}
\toprule
Model & Type & ROUGE-1 & ROUGE-2 & ROUGE-L & \bertscore \\ \midrule
\bertbase & \mask   &   5.33  &    0.72     &  4.94   &  -0.39$^*$  \\
\bertlarge &  \mask    & 8.05          &  0.63   &  7.85 & -0.27$^*$ \\ \midrule
\tfivebase & \template   &  14.00   &     2.71    & 12.58 & 2.2  \\
\tfivebase &  \ourmodel    &    14.01   &     2.60    & 12.57 & 6.1 \\ \midrule
\tfivelarge & \template   &  13.74   & 3.11     & 13.74 & 4.8\\
\tfivelarge &  \ourmodel    &   16.74   &   4.33 & 15.37 & 6.7 \\ \midrule
\bartbase & \template   &  17.17    & 5.60    & 16.32  &    3.9  \\
 \bartbase &  \ourmodel    &     18.89    &  5.87  &  17.96 & 6.3 \\ \midrule
\bartlarge & \template   &   19.54  &  7.57       &    18.49    & 7.0 \\
\bartlarge &  \ourmodel    &  20.58       & 7.32   & 19.58 & 7.6 \\ 
\bottomrule
\end{tabular}
\caption{Overview of the results compared to baselines. The table shows that \bartbase performs better than \tfivebase model and \bartlarge outperforms both. Both in terms of \rouge and \bertscore, we also observe that our \prompt approach outperforms \template approach. $^*$ - a negative score in \bertscore implies that the reference was dissimilar to the generated output. All experiments were done with 5 seeds, and reported are the average.}
\label{tab:results}
\end{table*}

In this section, we describe the experimental setup, and baselines for our approach. Since our \ourmodel approach is agnostic to the pretrained encoder-decoder architecture type, we perform experiments on two state-of-the-art \plm models - \bart and \tfive. 

\subsection{Experimental Setup}

\paragraph{Metrics}: We use the following evaluation metrics for evaluation for the \approach task: (i) \rouge \citep{lin2004rouge} and (ii) \bertscore \citep{Zhang2019BERTScoreET}. N-gram metrics such as \rouge are known to be limited, specifically for reasoning tasks. To mitigate this, we use \bertscore, which uses the similarity score between the reference and generated output using conceptual embeddings from \texttt{BERT} \citep{devlin-etal-2019-bert} model, which correlates better towards human judgements.

To perform the evaluation, we compare the generated sentence for the template against the gold annotations in the dataset.
We remove the template words from the output and only compare the slot filler concepts to avoid score inflation due to copying.
All the experiments were performed on a cluster of 8 NVIDIA V100 GPUs for about 32 GPU hours.

\subsection{Models} 

We follow the same experimental settings across the baseline and our approach for all the models. We initialize all the models with their pretrained weights.
We use commonly used encoder-decoder architectures for our experiments - \bartbase, \bartlarge, \tfivebase and \tfivelarge. 
The model settings are given below: 

\begin{itemize}
    \item \bartbase : This pretrained encoder-decoder transformer architecture is based on \citet{lewis-etal-2020-bart}. It consists of 12 transformer layers each with 768 hidden size, 16 attention heads and overall with 139M params. 
    \item \bartlarge : Larger version of \bartbase, with 24 transformer layers, 1024 hidden size, 16 heads and 406M params. 
    \item \tfivebase : The T5 model is also a transformer encoder-decoder model based on \citet{raffel-JMLR:v21:20-074} with 220M parameters with 12-layers each with 768 hidden-state, 3072 feed-forward hidden-state and 12 attention heads. 
    \item \tfivelarge : T5-Large model version comprises of 770M parameters with 24-layers with 1024 hidden-state, 4096 feed-forward hidden-state and 16 attention heads 
    \footnote{
Implementation adapted from Huggingface \citep{wolf-etal-2020-transformers} }. 
\end{itemize}

\subsection{Baseline Methods} 

\begin{table*}[t]
\centering
\small
\begin{tabular}{@{}lll@{}}
\toprule
Baseline & Template  & Output                                                    \\ \midrule
 \begin{tabular}[c]{@{}l@{}} \textsc{BERT} \mask \\ \citep{devlin-etal-2019-bert} \end{tabular} & \begin{tabular}[c]{@{}l@{}} \activity{\mask} has a  \qual{\mask} risk of \\ \activity{\mask} because \reason{\mask} \end{tabular} & \begin{tabular}[c]{@{}l@{}}Person who lives in a city \\ has a higher risk of depression \\  because of stress due to noise \end{tabular} \\ \midrule 
 \begin{tabular}[c]{@{}l@{}} \template \\ \citep{Donahue2020EnablingLM} \end{tabular} &  \begin{tabular}[c]{@{}l@{}} \activity{[S]person\_at\_location[/S]} has a \\  \qual{[S]higher/lower[/S]} risk of \\ \activity{[S]disease[/S]} because \\ \reason{[S]reason\_for\_risk[/S]} \end{tabular} & \begin{tabular}[c]{@{}l@{}}Person who lives in a city \\ has a higher risk of depression \\  because of stress due to noise \end{tabular}   \\ 
\bottomrule
\end{tabular}
\caption{Task Setup for baselines. In the first baseline, we query the \bert MLM model to check if vannila MLM models can solve the \approach task. In our second baseline, we use special tokens to indicate the start and end of each slot. In both the case, the models is trained to predict the output, which is a valid expansion for the template.}
\label{tab:baseline_setup}
\end{table*}

\begin{itemize}
    \item \textsc{BERT} \mask: To understand whether pretrained models contain the knowledge already, we try a masked language  modeling baseline \citep{devlin-etal-2019-bert} where we query the template using \mask tokens\footnote{Since mask tokens in \textsc{BERT} needs to be predetermined for this experiment, we try different variations with number of \mask tokens and report the best results.}. 
    
    \item \template : In this approach, we use the special token approach (\template) \citep{Donahue2020EnablingLM}, where we indicate the start and end of each template slot in the input and generate the output sentence
    
\end{itemize}

Table \ref{tab:baseline_setup} shows the baseline setup of the models for our task with a corresponding example.

\subsection{Results}

The results across various pretrained encoder-decoder approaches are shown in table \ref{tab:results}. In this table, we see that on average, \bart models perform better than \tfive models on average. We hypothesize this might be an effect of their pretraining task choices and corresponding datasets. We also observe that \prompt based models outperform the \template based approach. For all of the models and baselines, we used the greedy decoding strategy. 

Firstly, we find that \mask approach does not perform competitively compared to fine-tuning, showing that pretrained models are not easily amenable towards \approach without finetuning.  
Across all the experiments, we found that the \prompt approach outperforms \template approach across both \rouge and \bertscore scores for all models.   

\section{Factual Correctness Evaluation}

To further assess the quality of generated output, we perform additional factuality evaluation towards our best performing models - \template and \ourmodel approach using \bartlarge. Towards this we use the \factcc factuality metric \citep{kryscinski-etal-2020-evaluating}, which uses entailment classification to predict a binary factuality label between the source document and generated output.

Computing factuality using \factcc metric requires an input source document; (i.e.) the generated output is compared against the source document for factual correctness. For this evaluation setup, we augmented each generated output $y$ with a source document. Towards this, we use a large scale retrieval corpus based on \citet{nguyen2016ms}, and retrieve the top similar document $D$ \citep{Lin_etal_SIGIR2021_Pyserini} to a generated template expansion. Using the $(D, y)$ pairs, we compute the factual correctness of our best performing models. From the table \ref{tab:factuality_results}, we observe that our \ourmodel approach outperforms the \template approach for factual correctness by $\sim$14 points in accuracy. 


Additionally, we also perform human evaluation of factual correctness. For this experiment, three human judges annotated 100 unique samples for \textit{correctness} - that indicates how many samples were correct from a human perspective. 
We used our best performing \bartbase-\ourmodel model for this evaluation. In this experiment, a sentence generated by the model for a given template was given to each human judge and they were asked to evaluate whether the sentence was correct, given the template. 
The inter-annotator agreement on graph correctness was substantial with a Fleiss' Kappa score~\cite{fleiss1973equivalence} of 0.73.
From our evaluation, we found that human judges rated about 69\% of the sentences to be correct given a template, comparable to our \factcc evaluation metric numbers.
Both the automated and human evaluation suggests that our \ourmodel approach has better factual consistency.

\begin{table}[]
\centering
\begin{tabular}{@{}lll@{}}
\toprule
Model & Type & \factcc \\ \midrule
\bartlarge & \template   &   65.27       \\
\bartlarge &  \prompt    &  79.88     \\ 
\bottomrule
\end{tabular}
\caption{Factual consistency results. In this experiment, we show that our \ourmodel approach outperforms the \template approach in terms of factuality metric 
\factcc, showing its relative effectiveness}
\label{tab:factuality_results}
\end{table}



\section{Error Analysis} 

In this section, we analyze in detail how well language models perform template-expansion task for multihop reasoning. To understand the errors in depth, we complement our automated evaluation with manual error analysis. 
For this analysis, we randomly select 100 samples from the validation set predictions where the \rouge scores were low. We observe the following categories of errors that language models exhibit. Table \ref{tab:error_analysis} shows the common type of errors and a corresponding example for each type. 

\begin{table*}[!ht]
\centering
\resizebox{\textwidth}{!}{%
\begin{tabular}{@{}llll@{}}
\toprule
Error Type & Template & Gold Answer & Generated Answer \\ \midrule
\begin{tabular}[c]{@{}l@{}}Correct but \\ not in gold \end{tabular} & \begin{tabular}[c]{@{}l@{}}Children who are exposed to \\ \{environmental\_factor\} are often \\ at a higher  risk for \{disease\} \\ because \{reason\} \end{tabular} & \begin{tabular}[c]{@{}l@{}}Children exposed to second  \\ hand smoke  are at a  higher \\ risk for lung disease because \\ of breathing in the cigarette smoke\end{tabular} & \begin{tabular}[c]{@{}l@{}}Children who are exposed \\ to lead paint are often \\ at a higher risk for kidney failure \\ because lead causes kidney damage\end{tabular} \\ \midrule
\begin{tabular}[c]{@{}l@{}}Wrong \\ commonsense\\ concept\end{tabular} & \begin{tabular}[c]{@{}l@{}}People with \{certain\_socioeconomic \\ \_condition\}  are at higher risk of \\ \{disease\}  as they are more \\ exposed to \{reason\}\end{tabular} & \begin{tabular}[c]{@{}l@{}}Person who often inhale \\ a lot of dirt are \\ at a higher risk of hay fever \\ because of allergen content.\end{tabular} & \begin{tabular}[c]{@{}l@{}}Person who often take \\ less medications are at \\ a higher risk of hay fever \\ because of the drug ineffectiveness\end{tabular} \\ \midrule
\begin{tabular}[c]{@{}l@{}}Generic \\ Explanation\end{tabular} & \begin{tabular}[c]{@{}l@{}} When people with \\ \{certain\_co-morbidities\} show \\ \{symptoms\}, this is because \\ of \{reason\_for\_patient\_state\}\end{tabular} & \begin{tabular}[c]{@{}l@{}}When people with diabetes \\ show lethargy, this is \\ because of \\ high glucose levels.\end{tabular} & \begin{tabular}[c]{@{}l@{}}When people with heart \\ disease show  chest pain, \\ this is because \\ of the strain on the heart\end{tabular} \\ \midrule
\begin{tabular}[c]{@{}l@{}}Factually \\ Incorrect\end{tabular} & \begin{tabular}[c]{@{}l@{}}People with a \{health\_condition\} \\ should do \{an\_activity\}  \\ because \{reason\} \end{tabular} & \begin{tabular}[c]{@{}l@{}} People with a cardiovascular \\ disease should do exercise \\ since exercise burns \\  excess fat \end{tabular} & \begin{tabular}[c]{@{}l@{}} People with a flu diagnosis \\ should do exercise because \\ to stay active \end{tabular} \\ \bottomrule
\end{tabular}%
}
\caption{Error Analysis based on the \bartbase-\ourmodel model. We select 100 samples from the validation set and each row shows an example of each class of error. 
}
\label{tab:error_analysis}
\end{table*}

\paragraph{Error Type - Correct but not in gold (17\%) : }

In several cases, we observe that the output produced by the language models are correct despite not matching the gold answer. This phenomenon is evident when the input template contains multiple possible answers. While the gold answer in the example shown in Table \ref{tab:error_analysis} (first row) fills the template using \activity{smoking}, the model generates an answer related to \activity{kidney damage}. While correct, the automated generation metrics such as \rouge and \bertscore score such answers lower. 

\paragraph{Error Type - Wrong commonsense concept (8\%) :} 
In this category of error, the model generates the wrong specification for the given slot. For instance (second row in table \ref{tab:error_analysis}), the model mistakenly assumes \activity{person taking less medication} as a \activity{socioeconomic condition}. This error type gives a more nuanced understanding on which concept categories the model makes the most mistakes.  

\paragraph{Error Type - Generic Explanation (53\%): }
In several cases, the model resorts to generic explanation that are \textit{obvious}. A generic explanation repeats the same information as the rest of sentence as an explanation, thereby not providing any new information compared to the rest of the sentence. In the example shown in Table \ref{tab:error_analysis} (row 3), the explanation \reason{because of the strain of the heart} is already clear from the concept \activity{chest pain}. A generic explanation is often unreliable in explainable NLP systems since it does not provide any insight into the reasoning capability of the model \citep{https://doi.org/10.48550/arxiv.2205.03401}.

\paragraph{Error Type - Factually Incorrect (22\%) : }
Factual correctness is one of the biggest challenges in NLP applications \citep{petroni2020how,pagnoni-etal-2021-understanding}. The incorrect factual information is also acute for cross-domain reasoning applications as well. As shown in the example (row 4 in table \ref{tab:error_analysis}), the model incorrectly generates that \activity{people with flu diagnosis} should do \activity{exercise}. Factual correctness in generation models is an active area of research and we believe that template-based approaches can provide additional insight into this phenomenon. 

Overall, \approach remains a challenging task for \plm models, specifically on their factual correctness and we believe it opens several avenues for progress in this research direction. 

\section{Related Work} 

\paragraph{Knowledge Bases :} 
Knowledge Bases (KBs) have been the predominant approach to perform commonsense reasoning in the past \citep{Speer2013ConceptNet5A}. 
Some of the prominent knowledge bases for commonsense reasoning include  DBPedia \citep{mendes-etal-2012-dbpedia}, YAGO \citep{suchanek-yago} and NELL \citep{mitchell-nell} or extending KBs with domain knowledge \citep{khetan-etal-2022-mimicause}. In this work, we focus on \approach using LM, which can be viewed as a complementary using KBs for commonsense.  

\paragraph{Language Models for Generation based Reasoning:}
Using pretrained language models to generate knowledge has been studied for commonsense reasoning tasks.  \cite{sap2019atomic,Bosselut2019COMETCT,shwartz2020unsupervised,bosselut2021dynamic}.
Our work closely aligns with \citet{Bosselut2019COMETCT,bosselut2021dynamic}. 
Compared to \citet{Bosselut2019COMETCT}, where our goal is to extend towards more controllable commonsense reasoning. Our work is also related to recent chain-of-thought prompting approach \citep{dalvi-etal-2021-explaining,https://doi.org/10.48550/arxiv.2201.11903}, where a reasoning chain is first generated before the final solution. Compared to chain-of-thought prompting, our approach focuses on controllability of the reasoning process from input, via template slots.

\paragraph{Language Model Infilling :}
Our work also closely relates to the language model infilling work in the literature such as \citet{fedus2018maskgan} and \citet{Donahue2020EnablingLM}. Compared to these works which only look at cloze-test infilling, our work aims to expand templates that cannot be directly modeled as cloze-style. 

\section{Conclusion and Future Work}

In this paper, we present a novel \ourmodel approach that adapts language models to perform the \approach task by training them via prompting. We collect a dataset for the same, and show that such an approach allows higher control over the reasoning process by enabling practitioners to specify the nature of the template slots.
Through both automated and human metrics, we find that our \ourmodel approach outperforms the baselines while also maintaining high factuality. 
For future work, we hope to extend this line of work towards other controllable generation tasks such as story generation and summarization. 

\section{Acknowledgements}

We thank Byron Wallace for helpful discussions and providing valuable feedback for improving the work.

 

\bibliographystyle{acl_natbib}
\bibliography{acl}

\end{document}